# Cross-Lingual Word Alignment for ASEAN Languages with Contrastive Learning

Jingshen Zhang[1], Xinying Qiu[*1], Teng Shen[1], Wenyu Wang[3], Kailin Zhang[1], Wenhe Feng[2]
*[1]School of Information Science and Technolgy*
*[2]Laboratory of Language Engineering and Computing*
*Guangdong University of Foreign Studies, Guangzhou, China*
*[3]College of Computer and Software Engineering, Hohai University, Nanjing, China*
audbut0702@163.com, xy.qiu@foxmail.com, ErikGDUFS@gmail.com,
wangwyhhu@163.com, kailinzhang2022@gmail.com, wenhefeng@gdufs.edu.cn

*Abstract*— Cross-lingual word alignment plays a crucial role in various natural language processing tasks, particularly for low-resource languages. Recent study proposes a BiLSTM-based encoder-decoder model that outperforms pre-trained language models in low-resource settings. However, their model only considers the similarity of word embedding spaces and does not explicitly model the differences between word embeddings. To address this limitation, we propose incorporating contrastive learning into the BiLSTM-based encoder-decoder framework. Our approach introduces a multi-view negative sampling strategy to learn the differences between word pairs in the shared cross-lingual embedding space. We evaluate our model on five bilingual aligned datasets spanning four ASEAN languages: Lao, Vietnamese, Thai, and Indonesian. Experimental results demonstrate that integrating contrastive learning consistently improves word alignment accuracy across all datasets, confirming the effectiveness of the proposed method in low-resource scenarios. We will release our data set and code to support future research on ASEAN or more low-resource word alignment.

## I. Introduction

Cross-lingual word alignment is a fundamental task in natural language processing that aims to identify word-level correspondences between parallel sentences in two languages. Table I gives an example of word alignment. It plays a critical role in various downstream applications, such as machine translation [4][3], bilingual lexicon induction [20], and cross-lingual transfer learning [2]. However, word alignment remains challenging for low-resource languages due to the scarcity of parallel training data.

Recent work has focused on learning cross-lingual word embeddings to facilitate word alignment in low-resource settings. [14] proposed a BiLSTM-based encoder-decoder model that leverages both parallel sentences and subword information to learn robust cross-lingual embeddings without relying on large-scale pretraining. While their approach outperforms prior methods, it does not explicitly model the relationships between words in the embedding space.

TABLE I. AN EXAMPLE OF WORD ALIGNMENT

| | |
|---|---|
| **Input** | **Chinese:** 我是一个生态学家，我研究复杂性 <br> **Lao:** tôi là một nhà sinh thái học và tôi nghiên cứu sự phức tạp |
| **Output** | (tôi, 我), (là, 是), (một, 一个), (nhà sinh thái học, 生态学家), (nghiên cứu, 研究), (sự phức tạp, 复杂性) |

In this paper, we propose a novel contrastive learning framework for low-resource word alignment that addresses this limitation. Contrastive learning has recently emerged as a powerful technique for learning discriminative representations by pulling semantically similar examples closer and pushing dissimilar ones apart [22]. It has achieved state-of-the-art results in various fields, including computer vision [23], natural language processing [24], and speech recognition [7]. While contrastive learning has been applied to some cross-lingual tasks, its potential for cross-lingual word alignment has not been fully explored, particularly in low-resource settings.

Our approach extends Wada et al.'s [14] model by introducing a contrastive loss that explicitly models the relationships between word pairs in the cross-lingual embedding space. Specifically, we treat word pairs that are translations of each other as positive examples and non-translation pairs as negative examples. By learning to contrast positive pairs against negative pairs, the model is encouraged to learn a more discriminative embedding space where words with similar meanings are clustered together, and those with different meanings are pushed apart.

We evaluate our approach on five low-resource language pairs spanning four Southeast Asian languages: Lao, Vietnamese, Thai, and Indonesian. These languages are spoken by over 250 million people but have limited parallel corpora and NLP resources available. Our experiments show that incorporating contrastive learning can consistently improve word alignment accuracy over the base model and several strong baselines across all language pairs.

The main contributions of this work are:

1. We propose a novel approach that incorporates contrastive learning into a state-of-the-art BiLSTM-based encoder-decoder model for cross-lingual word alignment.

2. We conduct extensive experiments on five low-resource Southeast Asian language pairs, demonstrating consistent improvements in word alignment accuracy over the base model and strong baselines.

## II. Related Work

Two main approaches have been proposed for learning cross-lingual word embeddings. The first approach involves learning a matrix to align pre-trained monolingual word embeddings [9][10]. This method takes advantage of existing monolingual embeddings and learns a linear transformation to map them into a shared cross-lingual space.

The second approach is to jointly train cross-lingual word embeddings in a shared space from scratch [11][12][13]. This approach directly optimizes the cross-lingual embeddings using parallel corpora or bilingual dictionaries, ensuring that translations are close to each other in the shared space.

---

[*] Corresponding author

Recently, Wada et al. [14] proposed a BiLSTM-based encoder-decoder model with attention that achieves strong performance on cross-lingual word alignment for three endangered languages. Their model leverages both parallel sentences and subword information to learn robust cross-lingual embeddings in low-resource scenarios.

Contrastive learning (CTL) has emerged as a powerful technique for unsupervised and self-supervised representation learning [1]. The core idea of CTL is to optimize the similarity between a given query and its matched key while pushing away randomly chosen negative keys. CTL has been successfully applied in various domains.

In the context of cross-lingual word alignment, Wei et al. [8] proposed HiCTL, a hierarchical contrastive learning framework that learns universal representations across languages at both the sentence and word levels. By contrasting aligned sentences and words against negative examples, their approach achieves improved performance on machine translation tasks.

III. METHODOLOGY

Building upon the success of Wada et al. [14] and inspired by the potential of contrastive learning, we propose incorporating a contrastive learning objective into the BiLSTM-based encoder-decoder model. Specifically, we introduce a multi-view contrastive loss that selects negative samples from different perspectives during training.

The proposed method combines the strengths of the BiLSTM-based encoder-decoder architecture and contrastive learning to learn more discriminative cross-lingual word embeddings. By explicitly contrasting positive word pairs against negative examples, we aim to improve the alignment accuracy, particularly in low-resource settings where parallel data is limited.

*A. Base Model Architecture*

Our base model follows the BiLSTM-based encoder-decoder architecture with attention proposed by Wada et al. [14], which achieved state-of-the-art results on low-resource cross-lingual word alignment without relying on large pre-trained language models. As shown in Fig 1, the architecture consists of:

1. A shared bidirectional LSTM encoder that generates contextual word embeddings for the source language sentence by concatenating forward and backward LSTM states.

2. Language-specific attention mechanisms that compute a weighted sum of the encoder states, allowing the decoders to focus on relevant source words. Attention weights are calculated as softmax-normalized dot products between the current decoder hidden state and each encoder state.

3. Language-specific unidirectional LSTM decoders that predict target sentence words conditioned on previous target words and the source context vector from attention. Separate decoders are used for each target language and direction to account for word order differences.

The encoder employs bi-directional LSTM $f$, which are shared among all languages:

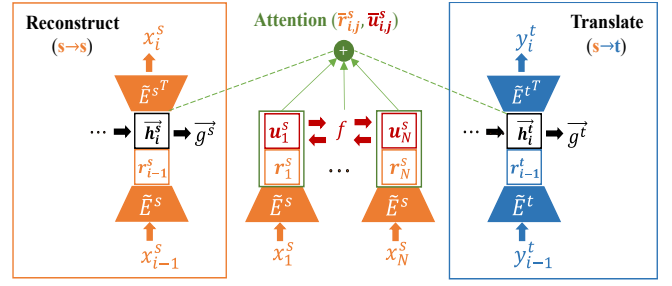

Fig 1. Model architecture of BiLSTM-based encoder-decoder with attention proposed by Wada et al., [14], which has been shown to be effective for low-resource word embedding learning.

$$r_i^s = E^s x_i^s \quad (1)$$

$$u_1^s, \dots, u_N^s = f(r_1^s, \dots, r_N^s) \quad (2)$$

where $x_i^s$ denotes a one-hot vector. In cross-lingual learning, we employ $r_i^s$ and $u_i^s$ as the static and contextualised word embeddings of $x_i^s$.

Given the encoder states $u^s$, the decoders $\vec{g}^t$ and $\overleftarrow{g}^t$ translate (when $s \neq t$) or reconstruct (when $s = t$) the input sentence left-to-right and right-to-left. Subsequently, we train separate decoders for each language and direction to allow for the differences of word order, which has been shown to effectively improving cross-lingual embeddings for distant languages. Therefore, decoding is performed independently in both directions:

$$r_i^t = E^t y_i^t \quad (3)$$

$$p(y_1^t \dots, y_M^t, EOS) = \prod_{i=1}^{M+1} p(y_i^t | \vec{h}_i, u^s) \quad (4)$$

$$\vec{h}_i = \vec{g}^t(\vec{h}_{i-1}, r_{i-1}^t) \quad (5)$$

$$p(BOS, y_1^t \dots, y_M^t) = \prod_{i=0}^{M} p(y_{M-i}^t | \overleftarrow{h}_{M-i}, u^s) \quad (6)$$

$$\overleftarrow{h}_i = \overleftarrow{g}^t(\overleftarrow{h}_{i+1}, r_{i+1}^t) \quad (7)$$

The output layer and attention mechanism are shared across the directions:

$$p(y_i^t | h_i, u^s, r^s) = softmax(E^{t^T} h_i') \quad (8)$$

$$h_i' = W(\bar{u}_i^s + \bar{r}_i^s + h_i)$$

$$\bar{u}_i^s = \sum_{j=1}^{N} \alpha_{i,j}^t u_j^s, \quad \bar{r}_i^s = \sum_{j=1}^{N} \alpha_{i,j}^t r_j^s \quad (9)$$

$$\alpha_{i,j}^t = \frac{exp(h_i u_j^s)}{\sum_{k=1}^{N} exp(h_i u_k^s)}$$

where $h_i$ denotes either $\vec{h}_i$ or $\overleftarrow{h}_i$, and $N$ is the number of words in the source sentence $x^s$. $E^t$ is the shared word embedding parameters for the output layer.

Additionally, subword embeddings is taken into account for incorporating orthographic information into word embeddings. For each word $w_i^\ell$, its subword-aware word embedding $\tilde{E}_{w_i}^\ell$ is shown as follows:

$$\tilde{E}^\ell_{w_i} = E^\ell_{w_i} + F(Z_{k \in Q(w_i)}) \quad (10)$$

where $F(\cdot)$ denotes the subword encoding function, average pooling is selected in our work; $Z_k$ denotes the $k$-th subword embedding and $Q(w_i)$ denotes the indices of the subwords included in $w_i$. For further details about this model, we refer the reader to Wada et al., [14].

### B. Constrastive Learning

**Original Contrastive Learning Loss**: Contrastive learning is a technique that aims to learn representations by contrasting positive pairs of similar examples against negative pairs of dissimilar examples. The key idea is to pull the representations of positive pairs closer together in the embedding space, while pushing apart the representations of negative pairs.

In the context of cross-lingual word alignment, contrastive learning can be used to refine the alignment of words across languages. By treating word pairs that are translations of each other as positive examples, and non-translation pairs as negative examples, contrastive learning can encourage the model to map words with similar meanings to be closer together in the shared cross-lingual embedding space, while separating words with different meanings.

Formally, the contrastive loss function, InfoNCE [7], is computed by measuring the similarity between a query $q$ and a key $k$ using a scoring function $\theta(q,k) = s(g(q), g(k))$:

$$\mathcal{L}_{constrastive} = -\log \frac{e^{\theta(q,k^+)/\tau}}{e^{\theta(q,k^+)/\tau} + \sum_i e^{\theta(q,k_i^-)/\tau}} \quad (11)$$

where $s(\cdot,\cdot)$ is the cosine similarity, $g_q(\cdot)$ and $g_k(\cdot)$ are non-linear projections for the query and key embeddings, $\tau$ is a temperature hyperparameter, $k^+$ is the positive key, and $k_1^-, k_2^-, \ldots$ are the negative keys.

**Encoder-Decoder Model with Contrastive Loss**: To apply contrastive learning to the BiLSTM encoder-decoder model, we consider the source sentence embedding as the query and the target sentence embedding as the positive key. The negative keys are sampled from the same training batch, as shown in Fig 2.

We represent a training batch of the source language sentences as $X = \{X_1, X_2, \ldots, X_n\}$, and its aligned target languages sentences as $Y = \{Y_1, Y_2, \ldots, Y_n\}$, where $n$ is the batch size. We firstly encode the source language sentence to obtain the embedding representations $\mathbb{M}_E(X_i)$, and we then extract its corresponding aligned target language embedding representations by $\mathbb{M}_D(Y_i)$. To encourage $\mathbb{M}$ to draw closer to the positive samples while away from the negative samples, we consider $Y_i$ as the positive corresponding to $X_i$.

**Negative Sampling for Translation and Reconstruction:** Please note that, negative sampling strategy for translation differs from that used for reconstruction. For reconstruction, we naturally define other instances in the same batch as negative samples (i.e., $X^x_{\setminus i}$, denoted as $\{k_1^-, k_2^-, \ldots, k_{n-1}^-\}$). As for translation, we adopt two negative sampling strategies:

1) **Intra-view**: defining non-corresponding translations in the same training batch as negative instances, i.e., $X^x_{\setminus i}$.

2) **Inter-view + Intra-view**: Similar to Zhu et al. [6] and Wei et al. [8], we select the negative instances from $X$ and $Y$, i.e., $X^x_{\setminus i} \cup Y^t_{\setminus i}$, denoted as $\{k_1^-, k_2^-, \ldots, k_{2n-2}^-\}$.

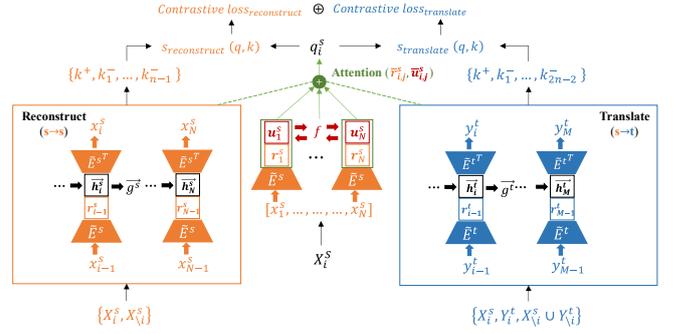

Fig 2. Our proposed model architecture.

**Contrastive Loss for Source Sentence**: The contrastive loss for $X_i$ of "Inter-view + Intra-view" is defined as:

$$\mathcal{L}_{ctl}(X_i) = -\log \frac{e^{\theta(q,k^+)/\tau}}{e^{\theta(q,k^+)/\tau} + \text{logit}_{inter} + \text{logit}_{intra} \cdot \mu} \quad (12)$$

$$logit_{inter} = \sum_{j=1}^{y \setminus i} e^{\theta(q,k_j^-)/\tau}, \quad logit_{intra} = \sum_{j=1}^{x \setminus i} e^{\theta(q,k_j^-)/\tau}$$

$$\mu = \begin{cases} 0, & inter\text{-}view \\ 1, & inter\text{-}view + intra\text{-}view \end{cases}$$

where the $q$ is denoted as $f(\mathbb{M}(X_i))$, $f(\cdot)$ is the aggregate function, in our work, we employ average pooling and max pooling respectively. $\mu$ represents the gate value for negative sampling strategy.

**Contrastive Loss for Target Sentence:** Symmetrically, we also expect the target $Y_i$, denoted as $\tilde{q}$, to be as similar as the source $X_i$, denoted as $\tilde{k}^+$, while dissimilar to all other negative instances in the same training batch, therefore:

$$\mathcal{L}_{ctl}(Y_i) \quad (13)$$
$$= -\log \frac{e^{\theta(\tilde{q},\tilde{k}^+)/\tau}}{e^{\theta(\tilde{q},\tilde{k}^+)/\tau} + \widetilde{logit_{inter}} + \widetilde{logit_{intra}} \cdot \mu}$$

$$\widetilde{logit_{inter}} = \sum_{j=1}^{y \setminus i} e^{\theta(\tilde{q},\tilde{k}_j^-)/\tau}, \quad \widetilde{logit_{intra}} = \sum_{j=1}^{x \setminus i} e^{\theta(\tilde{q},\tilde{k}_j^-)/\tau}$$

$\mu$ represents the gate value for negative sampling strategy.

**Contrastive Loss for Translation** is formulated as:

$$\mathcal{L}_{translation} = \frac{1}{2n} \sum_{i=1}^{n} \{\mathcal{L}_{ctl}(X_i) + \mathcal{L}_{ctl}(Y_i)\} \quad (14)$$

**Contrastive Loss for Reconstruction** is formulated as:

$$\mathcal{L}_{reconstruct} \quad (15)$$
$$= \frac{1}{n} \sum_{i=1}^{n} \left\{ -\log \frac{e^{\theta(q,k^+)/\tau}}{e^{\theta(q,k^+)/\tau} + logit_{intra}} \right\}$$

**Combined Loss for our proposed strategy** could be summarized as follows, where $s$ and $t$ refer to source and target sentence, respectively:

$$\mathcal{L}_{combine} \begin{cases} \mathcal{L}_{translation} = \frac{1}{2n}\sum_{i=1}^{n}\{\mathcal{L}_{ctl}(X_i) + \mathcal{L}_{ctl}(Y_i)\}, s = t \\ \mathcal{L}_{reconstruct} = \frac{1}{n}\sum_{i=1}^{n}\{\mathcal{L}_{ctl}(X_i)\}, s \neq t \end{cases}$$
(16)

## IV. EXPERIMENTS

### A. Data

We conduct experiments on real-world data sets in four ASSEN languages: Thai, Laotian, Vietnamese and Indonesian, specifically include Laotian-Chinese (Lao-Zh), Indonesian-Chinese (Id-Zh), Vietnamese-Chinese (Vi-Zh), Thai-Chinese (Th-Zh) and Laotian-Thai (Lao-Th).

We select OPUS[2] as our corpus source. Additionally, since the data sets of Lao-Zh and Lao-Th are extremely limited, we incorporate some bilingual alignment sentences from Glosbe[3]. Word dictionaries for Id-Zh, Vi-Zh, and Th-Zh are from Lingea[4], while Lao-Th is obtained by pivoting Lao-Zh and Th-Zh dictionaries on Zh. Pairwise words include polysemy. Table II gives statistical details and sources of our data.

TABLE II. DATASET STATISTICS

| Source - Target | Pairwise sentences Source (number) | Pairwise words Source (number) |
|---|---|---|
| Lao - Zh | OPUS + Glosbe (1503) | Glosbe (16712) |
| Id – Zh | OPUS (15000) | Lingea (7939) |
| Vi – Zh | OPUS (6916) | Lingea (4272) |
| Th – Zh | OPUS (50000) | Lingea (6595) |
| Lao - Th | OPUS + Glosbe (3190) | Lingea (2272) |

For gaining shared subword-embeddings, following Wada et al., [14], we employ SentencePiece for subword segmentation except Chinese. Because of the diversity of Chinese, we opt to uniformly segment Chinese words into character level.

### B. Baselines

We compare our model against a range of cross-lingual models include statistic-based, neural-based and PLM-based.

**Static PLMs**. We set up a PLMs-based benchmark model that aligns the words by extracting the static embeddings for each token. And we respectively select mBERT [16][5] and XLM-R [17][6].

**Sim-Align** [18]. A PLM-based word aligner without fine-tuning on any parallel data. In our work, we implement their IterMax model with default parameters and respectively extract static embeddings using mBERT and XLM-R.

**Fast-Align** [19]: A popular statistical word aligner which serves as a streamlined and an efficient reparameterization of IBM Model 2.

**GIZA++** [20]: An implementation of IBM models. Following previous study [5], we employ five iterations for each of Model 1, the HMM model, Model 3 and Model 4.

**Base Model** [14]. Following Wada, we train a BiLSTM-based encoder-decoder model with attention mechanism using parallel sentence pairs, treating it as the base model without contrastive learning loss.

### C. Evaluation Metric

We align each word in a sentence to the closest word in its translation. And we use cross-domain similarity local scaling [21] to calculate word similarity:

$$CSLS(x, y) = 2\cos(x, y) - \frac{1}{K}\sum_{y_t \in \mathcal{N}_T(x)}\cos(x, y_t) - \frac{1}{K}\sum_{x_t \in \mathcal{N}_S(y)}\cos(x_t, y)$$
(17)

where $\cos(x, y)$ denotes cosine similarity between $x$ and $y$ which are word embeddings, and $\mathcal{N}_T(x)$ and $\mathcal{N}_S(y)$ denote the $K$ nearest words to $x$ or $y$ in a target or source sentence; We set $K$ to 3 in the word alignment task. We extract the nearest word from the whole target vocabulary and verify whether they are listed as translations in the dictionary, and report P@1.

### D. Implementation Details

We set epoch to 200 for Lao-Zh and Id-Zh and 20 for the others. We use the Adam optimizer [15] with a learning rate of 2e-5 and a batch size of 16. Temperature hyper-parameter $\tau$ is set to 0.5. We use the Rectified Linear Unit (ReLU) as the activation function. We take the average over three runs as the final score.

TABLE III. MAIN RESULTS

| Method | Lao-Zh | Id-Zh | Vi-Zh | Th-Zh | Lao-Th |
|---|---|---|---|---|---|
| Static PLM (mBERT) (Devlin et al., 2019) | 1.35 | 36.08 | 17.44 | 3.59 | 4.12 |
| Static PLM (XLM-R) (Conneau et al., 2020) | 16.82 | 35.6 | 19.83 | 30.47 | 29.54 |
| Sim-Align (mBERT) (Sabet et al., 2020) | 3.38 | 38.88 | 21.02 | 7.27 | 9.19 |
| Sim-Align (XLM-R) (Sabet et al., 2020) | 17.31 | 40.36 | 23.61 | 35.54 | 37.09 |
| Fast-Align (Dyer et al., 2013) | 15.4 | 36 | 18.4 | 33.8 | 33.1 |
| GIZA++ (Och and Ney, 2003) | 16.31 | 67.51 | 20.14 | 52.35 | 44.28 |
| Base Model (Wada et al., 2021) | 54.4 | 73.536 | 44.66 | 62.434 | 69.266 |
| ***Ours*** | **56.53** | **74.114** | **45.657** | **62.573** | **69.878** |

---

[2] https://opus.nlpl.eu/index.php
[3] https://glosbe.com/
[4] https://www.dict.com/
[5] https://huggingface.co/google-bert/bert-base-multilingual-cased
[6] https://huggingface.co/FacebookAI/xlm-roberta-base

TABLE IV. STATISTICAL SIGNIFICANCE TESTS

| Methodology | Static PLM (mBERT) | Static PLM (XLM-R) | Sim-Align (mBERT) | Sim-Align (XLM-R) | Fast-Align | GIZA++ | Base Model |
|---|---|---|---|---|---|---|---|
| t-test | *0.002 | *0.0 | *0.003 | *0.0 | *0.0 | *0.023 | *0.058 |

## V. RESULTS & ANALYSIS

### A. Main Results

Table III reports the best performance of our proposed strategy compared against previous baselines. The best score for each language is in bold. Scores for the base model [14] and our model are averaged over three runs with different random seeds (0, 1, 2). The BiLSTM-based model with attention mechanism[14] outperforms a series of baseline models for ASEAN word alignment under low-resource scenarios, with an average improvement of 38.25 P@1 over the best-performing PLM-based method (SimAlign with XLM-R) across the five language pairs.

Incorporating contrastive learning loss further improves performance by an average of 0.75 P@1, achieving state-of-the-art results on all five datasets. Our model obtains the highest scores of P@1 on Lao-Zh, Id-Zh, Vi-Zh, Th-Zh, and Lao-Th respectively. This validates the **effectiveness of integrating contrastive learning for word alignment**. Table IV reports the statistical significance tests of our proposed strategy compared against multiple methods.

Pretrained language models (PLMs) perform poorly on ASEAN word alignment due to the lack of large-scale pretraining data for these low-resource languages. For example, mBERT achieves only 1.35 P@1 on Lao-Zh, 3.59 on Th-Zh, and 4.12 on Lao-Th, significantly underperforming statistical alignment methods.

Fig. 3 compares the score trends of our contrastive learning model versus the base model over three runs with different random seeds. The model with contrastive learning exhibits greater score variation.

For example, the base model for Id-Zh has a standard deviation of 0. However, the model incorporating contrastive learning has a standard deviation of 1.12. The larger variation is evident from the red shaded area being consistently wider than the green shaded area. We hypothesize that this is due to differences in similarity between the randomly sampled negative and positive pairs across runs. When negatives are similar but not identical to positives, the model performance can be improved by learning more nuanced differences. Conversely, highly dissimilar negatives may hinder the learning process.

### B. Ablation Studies

**Comparison with Base Model**: Table V presents detailed ablation study results. Bold indicates the best score for each language pair. ↑ and ↓ denote performance improvement and degradation, respectively, compared to the base model [14]. Avg.P. and Max.P. refer to average and max pooling aggregation.

In 4 out of 5 language pairs, integrating contrastive learning improves the base model's performance in at least 3 out of 4 metrics. The improvement is most significant on Vietnamese-Chinese (Vi-Zh), with gains of 0.528, 0.469, 0.997, and 0.352 (p-value < 0.05) under different settings. However, contrastive learning slightly degrades performance on Thai-Chinese (Th-Zh) in three out of four settings, possibly because Th-Zh has a larger dataset compared to other language pairs. The relative improvement from contrastive learning may be more pronounced in low-resource settings, where the model can benefit more from the additional training signal, while larger datasets may reduce the relative gains.

**Aggregate Function**: Unlike PLMs which extract sentence features from the [CLS] token, BiLSTM models typically use average or max pooling for aggregation. Results show that average pooling slightly outperforms max pooling overall, achieving the highest scores on 4 out of 5 language pairs.

**Negative Sampling Strategy**: We compare two negative sampling approaches: inter-view and inter+intra-view, to study the impact of negatives from translation v.s. reconstruction. The inter+intra-view strategy proves to be more effective, which is likely because the greater diversity of negative samples enables the model to better capture fine-grained differences. It achieves the best performance on 3 out of 5 language pairs.

Overall, the ablation studies confirm the benefit of contrastive learning for enhancing word alignment in low-resource settings.

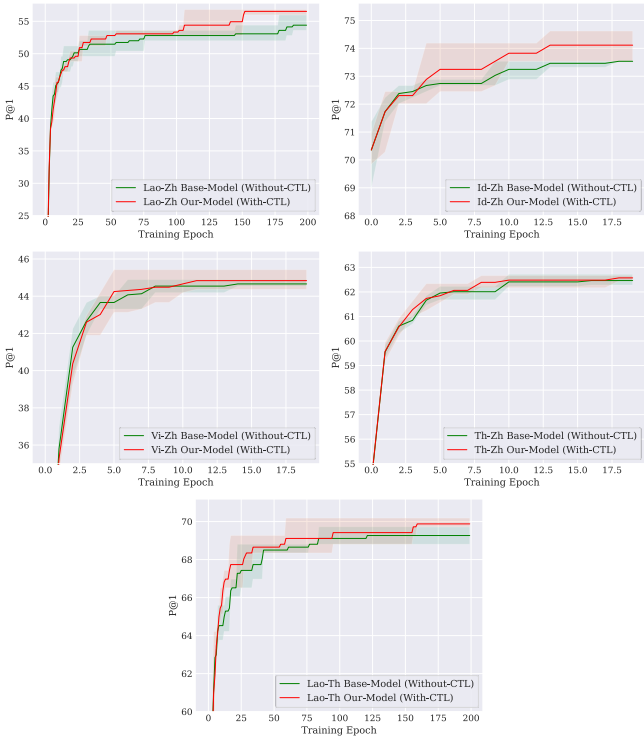

Fig 3. Comparison of model performance with (red) and without (green) contrastive learning over three runs with different random seeds. Solid lines show the average scores, while shaded areas indicate the performance range.

TABLE V. ABLATION STUDY RESULTS

| Language | Wada et al., 2021 | Incorporating Contrastive Learning | | | |
|---|---|---|---|---|---|
| | | Inter (view) | | Inter + Intra (view) | |
| | | Avg.P. | Max.P. | Avg.P. | Max.P. |
| Lao-Zh | 54.4 | **56.53↑** | 53.87↓ | 54.933↑ | 55.47↑ |
| Id-Zh | 73.536 | 73.753↑ | 74.042↑ | **74.114↑** | 73.536 |
| Vi-Zh | 44.66 | 45.188↑ | 45.129↑ | **45.657↑** | 45.012↑ |
| Th-Zh | 62.434 | 62.3↓ | **62.573↑** | 62.336↓ | 62.009↓ |
| Lao-Th | 69.286 | 68.96↓ | 69.725↑ | **69.878↑** | 69.878↑ |

## VI. CONCLUSION

In this work, we propose incorporating contrastive learning into a BiLSTM-based encoder-decoder model to improve cross-lingual word alignment for low-resource languages. By explicitly contrasting word pairs that are translations of each other against those that are not, our approach encourages the model to learn a more discriminative cross-lingual word embedding space. We evaluate our model on five language pairs covering four Southeast Asian languages: Lao, Vietnamese, Thai, and Indonesian.

Experimental results demonstrate that integrating contrastive learning consistently improves word alignment accuracy over the base model and several strong baselines, with an average gain of 0.75 P@1 across all five datasets. Ablation studies further show that using both inter-view and intra-view negative samples and average pooling for aggregation yields the best performance on most language pairs. Our work highlights the potential of contrastive learning to enhance cross-lingual word alignment in low-resource settings where parallel data is scarce. To support future research, we will release our code and datasets.

ACKNOWLEDGMENT

This work is partially supported by Guangzhou Science and Technology Plan Project (202201010729), and Guangdong Social Science Foundation Project (GD24CWY11). We thank the anonymous reviewers for their helpful comments and suggestions.

REFERENCES

[1] Nikunj Saunshi, Orestis Plevrakis, Sanjeev Arora, Mikhail Khodak, and Hrishikesh Khandeparkar. A theoretical analysis of contrastive unsupervised representation learning. In Proceedings of the 36th International Conference on Machine Learning, volume 97, pp. 5628–5637, Long Beach, California, USA, 09–15 Jun 2019. PMLR.

[2] Junjie Hu, Sebastian Ruder, Aditya Siddhant, Graham Neubig, Orhan Firat, and Melvin Johnson. 2020. Xtreme: A massively multilingual multi-task benchmark for evaluating cross-lingual generalization.

[3] Guillaume Lample, Alexis Conneau, Ludovic Denoyer, and Marc'Aurelio Ranzato. 2018. Unsupervised machine translation using monolingual corpora only. In *Proceedings of the International Conference on Machine Learning 2020* (pp. 4411-4421).

[4] Philipp Koehn. 2005. Europarl: A Parallel Corpus for Statistical Machine Translation. In Conference Proceedings: the tenth Machine Translation Summit, Phuket, Thailand. Asia-Pacific Association for Machine Translation.

[5] Thomas Zenkel, Joern Wuebker, and John DeNero. 2019. Adding interpretable attention to neural translation models improves word alignment. arXiv preprint arXiv:1901.11359.

[6] Yanqiao Zhu, Yichen Xu, Feng Yu, Qiang Liu, Shu Wu, and Liang Wang. 2020. Deep graph contrastive representation learning. arXiv preprint arXiv:2006.04131 (2020).

[7] Aaron van den Oord, Yazhe Li, and Oriol Vinyals. Representation learning with contrastive predictive coding. CoRR, abs/1807.03748, 2018.

[8] Xiangpeng Wei, Rongxiang Weng, Yue Hu, Luxi Xing, Heng Yu, and Weihua Luo. 2021. On learning universal representations across languages. *arXiv preprint arXiv:2007.15960.*.

[9] Chao Xing, Dong Wang, Chao Liu, and Yiye Lin. 2015. Normalized word embedding and orthogonal transform for bilingual word translation. In Proceedings of the 2015 NAACL, pages 1006–1011.

[10] Armand Joulin, Piotr Bojanowski, Tomas Mikolov, Hervé Jégou, and Edouard Grave. 2018. Loss in translation: Learning bilingual word mapping with a retrieval criterion. In Proceedings of the 2018 EMNLP, pages 2979–2984, Brussels, Belgium.

[11] Karl Moritz Hermann and Phil Blunsom. 2014. Multilingual models for compositional distributed semantics. In Proceedings of the 52nd Annual Meeting of the Association for Computational Linguistics (Volume 1: Long Papers), pages 58–68.

[12] Ali Sabet, Prakhar Gupta, Jean-Baptiste Cordonnier, Robert West, and Martin Jaggi. 2020. Robust crosslingual embeddings from parallel sentences. ArXiv 1912.12481.

[13] Efsun Sarioglu Kayi, Vishal Anand, and Smaranda Muresan. 2020. MultiSeg: Parallel data and subword information for learning bilingual embeddings in low resource scenarios. In Proceedings of the 1st Joint Workshop on Spoken Language Technologies for Under-resourced languages (SLTU) and Collaboration and Computing for Under-Resourced Languages (CCURL), pages 97–105.

[14] Takashi Wada, Tomoharu Iwata, Yuji Matsumoto, Timothy Baldwin, and Jey Han Lau. 2020. Learning contextualised cross-lingual word embeddings for extremely low-resource languages using parallel corpora. CoRR, abs/2010.14649.

[15] Diederik P. Kingma and Jimmy Ba. 2015. Adam: A method for stochastic optimization. In Proceedings of the 3rd International Conference on Learning Representations.

[16] Jacob Devlin, Ming-Wei Chang, Kenton Lee, and Kristina Toutanova. 2019. BERT: Pre-training of deep bidirectional transformers for language understanding. In Proceedings of the 2019 NAACL, pages 4171–4186, Minneapolis, Minnesota.

[17] Alexis Conneau, Kartikay Khandelwal, Naman Goyal, Vishrav Chaudhary, Guillaume Wenzek, Francisco Guzm´an, Edouard Grave, Myle Ott, Luke Zettlemoyer, and Veselin Stoyanov. Unsupervised cross-lingual representation learning at scale. In Proceedings of the 58th Annual Meeting of the Association for Computational Linguistics, pp. 8440–8451, Online, July 2020.

[18] Masoud Jalili Sabet, Philipp Dufter, François Yvon, and Hinrich Schütze. 2020. Simalign: High quality word alignments without parallel training data using static and contextualized embeddings. In Proceedings of the Conference on Empirical Methods in Natural Language Processing: Findings

[19] Chris Dyer, Victor Chahuneau, and Noah A. Smith. 2013. A simple, fast, and effective reparameterization of IBM model 2. In Proceedings of the 2013 NAACL, pages 644–648, Atlanta, Georgia. Association for Computational Linguistics.

[20] Franz Josef Och and Hermann Ney. 2003. A systematic comparison of various statistical alignment models. Computational Linguistics, 29(1):19–51.

[21] Alexis Conneau, Guillaume Lample, Marc'Aurelio Ranzato, Ludovic Denoyer, and Hervé Jégou. 2018. Word translation without parallel data. In Proceedings of the 6th International Conference on Learning Representations, Vancouver, BC, Canada.

[22] Hadsell, R., Chopra, S., & LeCun, Y. (2006). Dimensionality Reduction by Learning an Invariant Mapping. 2006 IEEE Computer Society Conference on Computer Vision and Pattern Recognition (CVPR'06), 2, 1735-1742.

[23] Chen, T., Kornblith, S., Norouzi, M., & Hinton, G. (2020, November). A simple framework for contrastive learning of visual representations. In International conference on machine learning (pp. 1597-1607). PMLR.

[24] Giorgi, J., Nitski, O., Wang, B., & Bader, G. (2020). Declutr: Deep contrastive learning for unsupervised textual representations. arXiv preprint arXiv:2006.03659.